\documentclass[runningheads]{llncs}

\usepackage{eccv}

\usepackage{eccvabbrv}

\usepackage{graphicx}
\usepackage{booktabs}
\usepackage{multirow}
\usepackage{mathabx}
\usepackage{array}
\usepackage{amsmath}
\usepackage[table]{xcolor}
\usepackage{enumitem}
\usepackage{subcaption}
\definecolor{top1}{RGB}{74,144,226}  % 深蓝(可读性友好)
\definecolor{top2}{RGB}{142,180,240} % 中蓝
\definecolor{top3}{RGB}{214,231,253} % 浅蓝
\DeclareMathOperator*{\argmax}{arg\,max}
\DeclareUnicodeCharacter{202C}{}

\usepackage[accsupp]{axessibility}  % Improves PDF readability for those with disabilities.

\usepackage{hyperref}

\usepackage{orcidlink}

\begin{document}

\title{Seeing What Matters: Lesion-Aware High-Resolution Patch Discovery and Fusion for Chest X-ray Report Generation}

\titlerunning{LePaX: Lesion-Aware High-Resolution Patch Discovery and Fusion}


\newcommand{\equalcontrib}{\textsuperscript{*}} 
\newcommand{\corrauth}{\textsuperscript{\dag}}

\author{
Yingshu Li\equalcontrib\inst{1}\orcidlink{0009-0009-4085-993X} \and
Yunyi Liu\equalcontrib\inst{1}\orcidlink{0009-0009-3157-895X} \and
Zhenghao Chen\inst{2}\orcidlink{0000-0003-0155-4462} \and
Tong Chen\inst{1}\orcidlink{0000-0003-4312-7151} \and
Zailong Chen\inst{3}\orcidlink{0009-0003-8431-5471} \and
Lingqiao Liu\inst{4}\orcidlink{0000-0003-3584-795X} \and
Lei Wang\inst{3}\orcidlink{0000-0002-0961-0441} \and
Luping Zhou\corrauth\inst{1}\orcidlink{0000-0001-8762-2424}
}

\authorrunning{Y. Li et al.}

\institute{
The University of Sydney,
Sydney, NSW 2006, Australia\\
\email{\{yingshu.li,yunyi.liu1,luping.zhou\}@sydney.edu.au,
tche2095@uni.sydney.edu.au}
\and
The University of Newcastle,
Newcastle, NSW 2308, Australia\\
\email{zhenghao.chen@newcastle.edu.au}
\and
University of Wollongong,
Wollongong, NSW 2522, Australia\\
\email{zc881@uowmail.edu.au, leiw@uow.edu.au}
\and
The University of Adelaide,
Adelaide, SA 5005, Australia\\
\email{lingqiao.liu@adelaide.edu.au}
}

\maketitle
\begingroup 
\renewcommand{\thefootnote}{} \footnotetext{\textsuperscript{*} Equal contribution. 
\textsuperscript{\dag} Corresponding author} 
\endgroup

\begin{abstract}
% Despite rapid advances in chest X-ray (CXR) foundation models, most radiology report generation (RRG) systems still rely on heavily downsampled inputs (e.g., $256{\times}256$) due to the fixed visual token budgets and architectural constraints of pretrained vision encoders. Such aggressive resizing suppresses subtle yet clinically important cues and limits the ability of current models to replicate real diagnostic workflows.
% Inspired by the diagnostic reasoning process of radiologists, who first localize suspicious regions and then perform detailed assessments, we formulate high-resolution perception in RRG as a constrained spatial resolution allocation problem, where the model must determine where to allocate additional high-resolution modeling capacity under a fixed token budget.
% To this end, We propose \textbf{Le}sion-\textbf{A}ware High-\textbf{R}esolution \textbf{Pa}tch Discovery and Fusion for Chest \textbf{X}-ray Reporting (LePaX). A Learnable Spatial Resolution Allocation (LSRA) module predicts a spatial utility map to select informative regions and extract high-resolution patches from native CXRs (up to $1920{\times}1920$). A Global–Regional Fusion (GRF) module then performs spatially aligned resolution write-back, injecting regional evidence into global representations without increasing the visual token budget.
% Extensive experiments on MIMIC-CXR, IU-Xray, and CheXpertPlus show that LePaX consistently surpasses previous state-of-the-art methods across both clinical metrics and linguistic metrics.

Despite rapid advances in chest X-ray (CXR) foundation models, most radiology report generation (RRG) systems still rely on heavily downsampled inputs (e.g., 256×256) due to the fixed visual token budgets of pretrained vision encoders, suppressing subtle yet clinically important cues present in native-resolution images. However, enabling high-resolution (high-res) perception remains challenging: naïve tiling causes prohibitive token inflation, while global compression suppresses subtle lesions and degrades diagnostic fidelity. Inspired by radiologists’ workflow, localizing suspicious regions before detailed high-res assessment. We propose \textbf{Le}sion-\textbf{A}ware High-\textbf{R}esolution \textbf{Pa}tch Discovery and Fusion for Chest \textbf{X}-ray Reporting (LePaX), the first RRG framework that enables efficient high-res CXR perception (up to 1920×1920) without increasing the vision-token count. LePaX formulates high-res perception as a constrained spatial resolution allocation problem under a fixed token budget and introduces two key components: Learnable Spatial Resolution Allocation (LSRA), which learns a spatial utility map that adaptively allocates limited high-res capacity to diagnostically relevant regions, enabling targeted extraction of high-res patches from native CXRs; and Global–Regional Fusion (GRF), which performs token-preserving region-to-global refinement by projecting high-resolution regional evidence back onto the global feature grid through spatially aligned resolution write-back, avoiding token inflation. Experiments on multiple CXR benchmarks demonstrate that LePaX consistently improves both clinical and linguistic metrics while enabling native-resolution CXR perception with over 10× fewer visual tokens than naïve high-res tiling.
\end{abstract}

\begin{figure}[h]
  \centering
   \includegraphics[width=\linewidth]{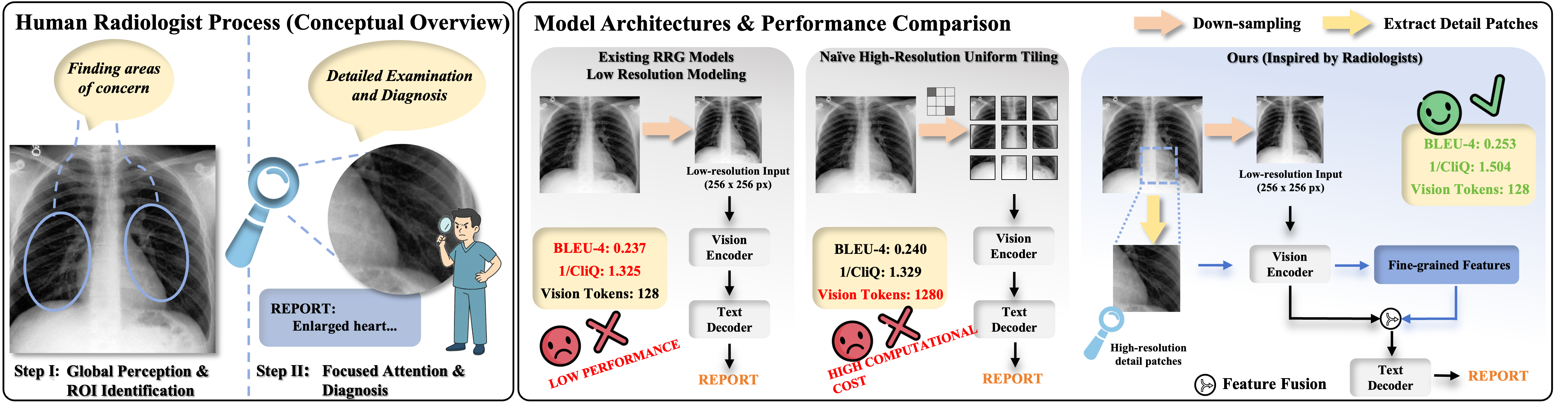}

   \caption{\textbf{Left}: Radiologists typically follow a two-stage workflow: a global survey to localize suspicious regions, followed by focused examination of these areas to verify specific findings (e.g., enlarged heart). \textbf{Right}: Conventional RRG models rely on fixed-size, low-resolution inputs and lack this global–local reasoning process. Our radiologist-inspired framework instead extracts and integrates high-resolution regional details, producing reports with improved clinical fidelity.}
   \label{fig:story}
\end{figure}
\section{Introduction}
\label{sec:intro}

Radiology report generation (RRG)\cite{Park_2025_CVPR, li2024kargen, huang2023kiut, yang2023radiology, liu2024multi, zhou2025review} aims to automatically produce clinically faithful diagnostic narratives and reduce radiologists’ workload. Recent advances in multimodal large language models (MLLMs)\cite{lillava, bai2025qwen2, xue2024xgen} have improved CXR report generation~\cite{wang2023r2gengpt, pellegrini2023radialog, liu2024bootstrapping, li2024kargen, wang2025cxpmrg, liu2026sat}. However, most RRG systems rely on heavily downsampled inputs (e.g., 256×256) to satisfy fixed-size encoder constraints. While resizing simplifies visual encoding, it suppresses subtle but clinically important cues present in native resolution CXRs (raw CXRs are multi-megapixel). This is problematic because diagnostically relevant abnormalities are often small, faint, and spatially sparse. Consequently, findings such as nodules, fine vascular markings, or faint consolidations may be attenuated, weakening lesion localization and increasing the risk of diagnostic omissions.

However, enabling high-resolution perception in MLLMs is computationally challenging. Transformer-based vision encoders represent images as visual tokens whose number grows with spatial resolution, increasing computational cost. A common solution is uniform tiling, where the image is divided into patches, but this strategy is poorly suited for medical imaging. Because abnormalities in CXRs are often sparse and anatomically localized, uniform tiling allocates many visual tokens to non-diagnostic regions, consuming token budgets without improving diagnostic sensitivity (Fig.~\ref{fig:story} right).

More fundamentally, such designs fail to reflect how radiologists interpret medical images. Clinical studies show that radiologists typically follow a global-to-local diagnostic workflow: they first perform a global survey to establish anatomical context and identify suspicious regions, and then zoom in for detailed verification~\cite{seah2021effect, brennan2018radiologists}. This workflow enables clinicians to allocate visual attention efficiently and integrate local evidence with global context. This raises a central question: \textit{How can radiology-aligned high-resolution perception be modeled efficiently within MLLMs so that systems better reflect real diagnostic workflows?}

Inspired by this diagnostic process, we argue that high-resolution perception in medical imaging should be selective rather than uniform. Instead of processing all regions at the same resolution, models should allocate high-resolution modeling capacity to diagnostically informative areas while preserving global contextual awareness. We therefore \textbf{formulate high-resolution perception as a constrained spatial resolution allocation problem}, where the model determines where allocating additional high-resolution modeling capacity yields the greatest diagnostic utility under a fixed visual token budget. By enabling non-uniform high-resolution reasoning, this formulation explicitly mirrors the radiologists’ global-to-local diagnostic strategy: global context is first established and then selectively refined with detailed local evidence.

Built on this formulation, we propose \textbf{LePaX} (Lesion-Aware High-Resolution Patch Discovery and Fusion), the first RRG framework enabling efficient high-resolution CXR perception (up to 1920×1920) without increasing visual token count. LePaX consists of two jointly optimized components: \textbf{Learnable Spatial Resolution Allocation (LSRA)} and \textbf{Global–Regional Fusion (GRF)}.

LSRA models the radiologists’ “where to zoom” decision. It learns a resolution allocation policy over the global feature grid by predicting a spatial utility map that indicates where additional high-resolution modeling capacity is most beneficial. Under a fixed token budget, LSRA selects a small number of high-resolution patches from the native-resolution image, enabling the model to preserve subtle pathological patterns and fine anatomical structures that would otherwise be suppressed by aggressive downsampling. During training, the allocation policy is jointly guided by weak localization priors (e.g., Grad-CAM~\cite{selvaraju2017grad} priors from disease classifiers) and the end-to-end report generation objective, allowing the model to learn diagnostically meaningful high-resolution allocation strategies.

However, lesion-centric regions alone are insufficient for reliable diagnosis. Radiologists also rely on \textbf{global image} to extract secondary cues (e.g., bilateral symmetry, cardiothoracic ratio) that inform diagnostic decisions~\cite{kundel2007holistic, gandomkar2021global}.
To emulate this holistic reasoning process, we introduce GRF, which models the integration of local evidence with global context. GRF GRF performs spatially grounded resolution write-back, injecting high-resolution regional features into corresponding locations within the global feature grid. This spatially grounded fusion enriches global representations with fine-grained diagnostic evidence while preserving anatomical coherence, all without increasing the visual token budget.

Our key contributions are summarized as follows:
\begin{itemize}
\item We present \textbf{LePaX}, a radiologist-inspired high-resolution RRG framework that formulates lesion-sensitive perception as a \emph{spatial resolution allocation} problem under fixed visual token budgets, explicitly reflecting the global-to-local diagnostic workflow.
\item We introduce two jointly optimized modules: (1) \textbf{LSRA}, which learns a \emph{resolution allocation policy} to decide where additional high-resolution capacity should be invested, and (2) \textbf{GRF}, a spatially grounded \emph{resolution write-back refinement} mechanism that injects high-resolution evidence into the global representation without token explosion.
\item Extensive experiments on \textbf{MIMIC-CXR}, \textbf{IU-Xray}, and \textbf{CheXpertPlus} demonstrate consistent gains on both NLG and clinical metrics.
\end{itemize}

% \newpage
\section{Related Work}
\label{sec:Related Work}

\subsection{Radiology Report Generation}
Radiology report generation (RRG) is a challenging task, and recent advances have improved performance~\cite{wang2023metransformer,hou2023organ,bu2024instance,huang2023kiut,Park_2025_CVPR}. Early methods mainly relied on encoder–decoder architectures, while more recent work adopts Transformer-based frameworks~\cite{vaswani2017attention}. METransformer~\cite{wang2023metransformer} introduces learnable expert tokens for multi-specialist reasoning. KiUT~\cite{huang2023kiut} incorporates structured medical knowledge into a U-shaped Transformer for disease-aware visual–text alignment. DART~\cite{Park_2025_CVPR} proposes a disease-aware alignment framework with a self-correcting re-alignment mechanism to improve factual consistency and reduce hallucinations.

The emergence of multimodal large language models (MLLMs)\cite{xue2024xgen,lillava,bai2025qwen2} has shifted RRG from traditional encoder–decoder paradigms toward unified vision–language generation\cite{li2024kargen,pellegrini2023radialog,li2023comprehensive,liu2024bootstrapping,li2023llava,chen2025enhancing,wang2025cxpmrg}. R2GenGPT~\cite{wang2023r2gengpt} was among the first to leverage frozen LLMs with efficient visual–language alignment, achieving strong linguistic fluency with minimal training cost. LLaVA-Med~\cite{li2023llava} extends instruction-tuned LLMs to medical imaging via staged multimodal alignment and instruction tuning. Multi-Phased Supervision~\cite{chen2025enhancing} adopts curriculum-style training from disease labels to entity–relation reasoning and full report generation. Most recently, MambaXray-VL~\cite{wang2025cxpmrg} combines autoregressive visual pretraining and image–text contrastive alignment within an LLaMA-based decoder, achieving strong fluency and clinically grounded reports.

Despite these advances, most frameworks operate on low-resolution CXRs, treating visual tokens as uniformly informative and overlooking the global-to-local reasoning radiologists use to link sparse abnormalities with textual evidence. This limitation motivates our \textbf{Lesion-Aware High-Resolution Patch Discovery and Fusion for Chest X-ray Report Generation (LePaX)}, a high-resolution yet efficient framework that preserves fine-grained anatomical details while maintaining global coherence.

\subsection{High-resolution Multimodal Large Language Models}
Recent multimodal large language models (MLLMs) process high-resolution or irregularly shaped images through tiling, cropping, or feature compression to balance spatial fidelity and efficiency~\cite{lillava,bai2025qwen2,chen2024far,zhu2023minigpt,ye2024mplug}. LLaVA-OneVision~\cite{lillava} adopts the Higher AnyRes scheme with multi-crop inputs and interpolation to maintain token budgets; InternVL-1.5~\cite{chen2024far} partitions images into 448×448 tiles with a global thumbnail and compresses features via pixel-shuffle; Qwen2.5-VL~\cite{bai2025qwen2} employs dynamic-resolution ViTs with windowed attention and token merging for efficient native-scale encoding.
Although effective for general MLLMs, these tiling and compression strategies often average fine-grained cues, weakening lesion localization and clinical grounding in radiology report generation (RRG). Unlike natural images, where dense spatial coverage is beneficial, medical imaging features \textit{spatially sparse, anatomically localized}, and clinically weighted abnormalities. Uniform tiling or global merging thus introduces redundant, non-diagnostic tokens and suppresses disease-relevant evidence.
In contrast, \textbf{LePaX} formulates high-resolution modeling as a \textbf{learnable spatial resolution allocation (LSRA)} problem. A utility map predicts informative regions where high-resolution patches are allocated, and the resulting regional features are written back to the global grid via a spatially grounded global–regional fusion (GRF) module, enabling detailed lesion modeling without increasing the token budget.

\begin{figure}[t]
  \centering
   \includegraphics[width=\linewidth]{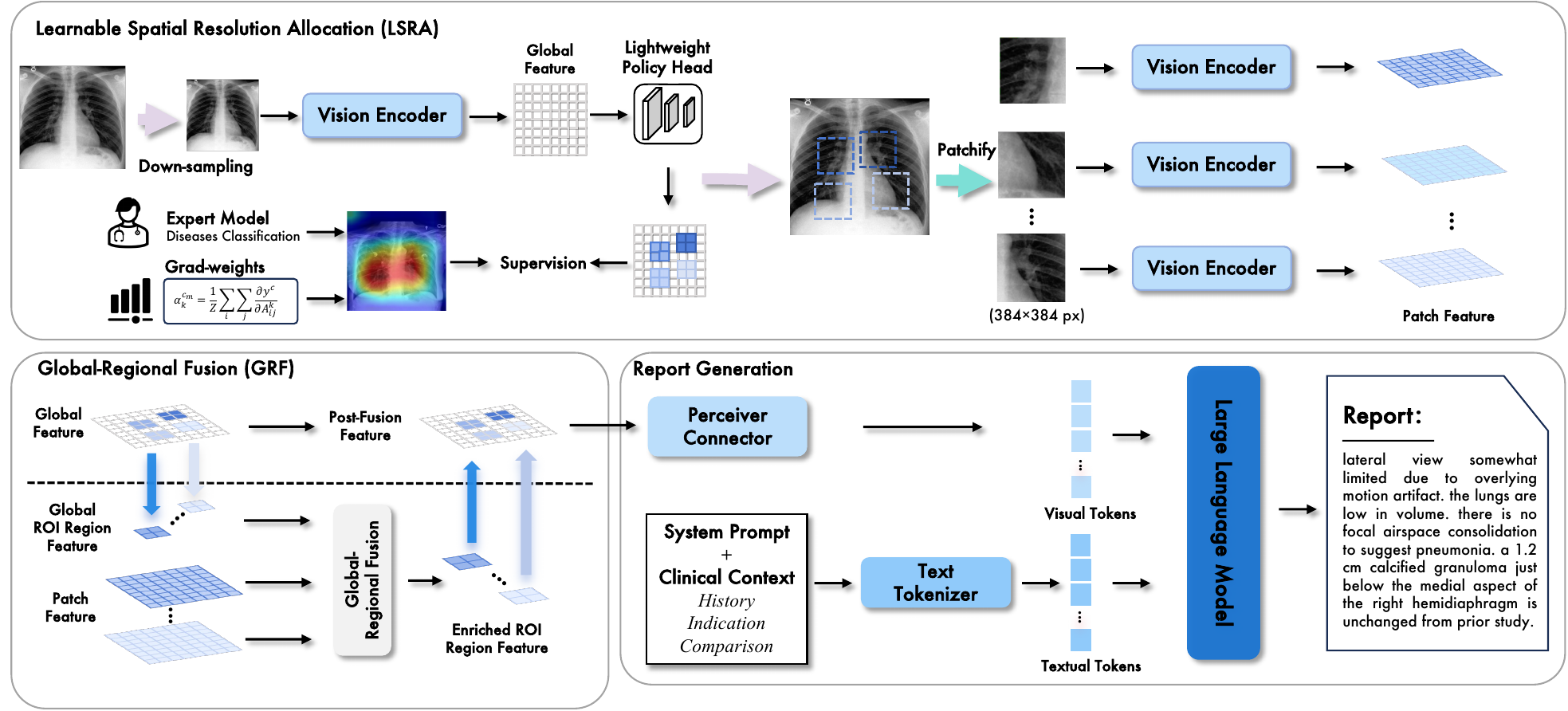}
   \caption{Overview of the proposed \textbf{LePaX} framework. A lightweight policy head predicts a spatial utility map from global features to allocate a fixed number of high-resolution patches under a resolution budget (LSRA). Global and patch features share a vision encoder and are fused via utility-guided Global–Regional Fusion (GRF), which writes high-resolution evidence back to the global feature grid. The fused visual tokens are compressed by a Perceiver connector and passed to an LLM conditioned on clinical prompts to generate grounded radiology reports.
    }
   \label{fig:overall}
\end{figure}
\section{Methodology}
\label{sec:Methodology}

\subsection{Overall Framework}
As illustrated in Fig.~\ref{fig:overall}, \textbf{LePaX} is a lesion-aware radiology report generation framework that enables adaptive high-resolution perception under a fixed visual token budget. Rather than uniformly processing high-resolution images, LePaX allocates limited high-resolution capacity to diagnostically informative regions. The framework introduces two key mechanisms.
\textbf{(1) Learnable Spatial Resolution Allocation (LSRA).}
LSRA predicts a grid-level \emph{resolution utility map} from global visual features, indicating where additional high-resolution modeling capacity should be allocated. 
Under a fixed token budget, LSRA selects a small set of spatial locations and extracts high-resolution patches from high-resolution CXRs (up to $1920{\times}1920$ px), preserving subtle pathological cues while avoiding uniform high-resolution processing.
\textbf{(2) Global--Regional Fusion (GRF).}
The global image and selected patches are encoded by a shared backbone to ensure geometric alignment in feature space. 
GRF performs region-conditioned update within the selected regions of interest (ROIs), injecting high-resolution evidence back into their spatially corresponding global tokens without increasing the visual token budget. 
This produces fused visual representations
$F_g^{\text{fused}} \in \mathbb{R}^{N_v \times D_v}$.
The fused representation is then processed by a Perceiver connector and used to condition the LLM for radiology report generation.

\noindent\textbf{Visual Encoder.}~~
The global image and selected high-resolution patches are encoded by a shared vision backbone $\phi_{\text{img}}$ to maintain geometric alignment in feature space:
\begin{equation}
F = \phi_{\text{img}}(X) \in \mathbb{R}^{N_v \times D_v}.
\end{equation}
We denote the global tokens as $F_g$ and the $k$-th patch tokens as $F_p^k$.

\noindent\textbf{Perceiver Connector.}~~
To align fused visual features with the language model, we employ a Perceiver connector that compresses $F_g^{\text{fused}}$:
\begin{equation}
\begin{aligned}
F_c &= \mathrm{Perceiver}(F_g^{\text{fused}}) \\
    &= \mathrm{CrossAttn}\!\big(Q_0,\; K{=}[F_g^{\text{fused}}; Q_0],\; V{=}[F_g^{\text{fused}}; Q_0]\big)
       \; \in \mathbb{R}^{N_c \times D_c},\; N_c \ll N_v .
\end{aligned}
\end{equation}
where $Q_0 \in \mathbb{R}^{N_c \times D_c}$ denotes learnable latent tokens and
$[\cdot\,;\cdot]$ denotes token-wise concatenation.

\noindent\textbf{Text Decoder.}~~
Conditioned on $F_c$ and textual prompt $C$, the LLM autoregressively generates the radiology report
$Y=\{y_1,\ldots,y_T\}$:
\begin{equation}
p(y_t \mid F_c, C, y_{<t}) = \mathrm{LLM}(F_c, C, y_{<t}).
\end{equation}

\subsection{Resolution Allocation Formulation}

Let $\mathbf{X} \in \mathbb{R}^{H_0 \times W_0 \times 3}$ denote a high-resolution chest X-ray and 
$\mathbf{X}_{\text{low}}$ its downsampled version used for policy prediction. 
Given a fixed patch budget $B$, we formulate report generation as a constrained spatial resolution allocation problem that assigns limited high-resolution modeling capacity to the most informative regions.
We define a learnable allocation policy $\pi_\theta$ that selects spatial locations
$\mathcal{S} = \{(i_k,j_k)\}_{k=1}^{K}$ with $K \le B$.
The objective is
\begin{equation}
\max_{\theta}
\;\mathbb{E}_{(\mathbf{X},Y)\sim\mathcal{D}}
\left[
\log p_\phi\!\left(
Y \mid \mathbf{X}_{\text{low}}, \pi_\theta(\mathbf{X}_{\text{low}}, \mathbf{X})
\right)
\right]
\quad
\text{s.t.}
\quad |\pi_\theta(\mathbf{X}_{\text{low}}, \mathbf{X})| \le B .
\end{equation}
Here $\pi_\theta(\mathbf{X}_{\text{low}}, \mathbf{X})$ predicts spatial locations from $\mathbf{X}_{\text{low}}$ and crops the corresponding high-resolution patches from $\mathbf{X}$.
LSRA parameterizes the allocation policy $\pi_\theta$, while GRF integrates the selected high-resolution evidence with global representations without increasing the number of visual tokens.

\subsection{Learnable Spatial Resolution Allocation (LSRA)}

High-resolution CXRs contain subtle pathological cues, yet most vision encoders operate on low-resolution inputs, attenuating fine-grained diagnostic patterns. Meanwhile, clinically relevant findings are often spatially sparse, making uniform high-resolution processing computationally inefficient.  
To address this, LSRA learns a spatial utility function over the global feature grid, enabling adaptive high-resolution allocation under a fixed resolution budget.

\noindent\textbf{Resolution Utility Learning.}~~
Let $F_g \in \mathbb{R}^{C \times H \times W}$ denote the global feature extracted from the low-resolution CXR. LSRA predicts a spatial allocation map
\begin{equation}
Z = g(F_g) \in \mathbb{R}^{1 \times H \times W}, 
\qquad
U = \sigma(Z),
\end{equation}
where $U_{ij}$ represents the utility of allocating additional modeling capacity to location $(i,j)$.

\noindent\textbf{Disease-Aware Spatial Prior.}~~
Learning a spatial allocation policy from report supervision alone can be unstable during early optimization. 
Since most medical datasets lack pixel-level lesion annotations, we introduce a disease-aware spatial bias by deriving a coarse prior using Grad-CAM from a pretrained multi-label disease classifier $f_{\text{cls}}$. Given an input image $X$, let $A \in \mathbb{R}^{C_a \times H_0 \times W_0}$ denote the final convolutional feature map and $y^c$ the logit of class $c$. The Grad-CAM activation for class $c$ is

\begin{equation}
H^{c}
=
\mathrm{ReLU}
\left(
\sum_{k=1}^{C_a}
\alpha_k^c A^k
\right),
\quad
\alpha_k^c
=
\frac{1}{H_0W_0}
\sum_{i,j}
\frac{\partial y^c}{\partial A^k_{ij}} .
\end{equation}

Class-wise activation maps are aggregated and normalized to obtain a spatial prior $M^{\text{cam}}$, providing weak spatial guidance for the allocation policy. The report generation loss further refines the utility toward task-relevant regions. We implement this prior as a regularization term on the allocation logits:

\begin{equation}
\mathcal{L}_{\text{prior}}
=
\mathrm{BCEWithLogits}(Z, M^{\text{cam}}).
\label{eq:policy}
\end{equation}

This prior is used only to regularize LSRA during training; at inference, patch locations are selected solely from the learned utility map $U$, without computing Grad-CAM or invoking any external disease classifier.

\noindent\textbf{Allocation Strategy.}~~
Given the utility map $U=\sigma(Z)$, LSRA allocates a fixed high-resolution budget by selecting the top-$K$ spatial locations on the grid.

\noindent\textit{Training.}~~
To enable gradient-based optimization of the allocation policy, we employ a Gumbel-Top-$K$ relaxation with a straight-through estimator (STE). 
During training, Gumbel noise perturbs the allocation logits to produce stochastic rankings, from which a hard Top-$K$ mask is obtained for patch extraction.
$\mathcal{L}_{\text{policy}}$ provides a spatial prior for the allocation logits, while $\mathcal{L}_{\text{RRG}}$ (defined in Eq.~\ref{eq:loss}) supplies additional task-driven signals, encouraging the learned utility map to prioritize regions beneficial for report generation.

\noindent\textit{Inference.}~~
At inference time, LSRA selects the $K$ spatial locations with the highest utilities.
To encourage spatial diversity and reduce redundant overlap, we apply an NMS-like constraint that enforces a minimum grid distance $d_{\min}$:
\begin{equation}
\mathcal{S}=\{(i_k,j_k)\}_{k=1}^{K}
=\mathrm{TopK\!+\!NMS}(U;K,d_{\min}).
\end{equation}

Each selected location defines a region-of-interest (ROI) $\Omega_k$ on the global feature grid.
The ROI center is projected to the original image to obtain pixel coordinates $(x_k,y_k)$, from which a fixed-size patch is extracted:
\begin{equation}
X_k=\mathrm{Crop}\big(X;x_k,y_k,p_w,p_h\big),
\qquad
p_w=p_h=384.
\end{equation}

We emphasize that LSRA is not a Grad-CAM-based crop selector: 
$M^{\text{cam}}$ only regularizes allocation learning during training, 
whereas inference selects patches solely from the learned utility map 
$U=\sigma(Z)$ without Grad-CAM or external classifiers. 
Thus, LSRA acts as an end-to-end adaptive resolution allocation policy 
for report generation.

\subsection{Global-Regional Fusion (GRF)}

The global image and selected patches are encoded by a shared backbone, producing 
$F_g$ and $\{F_p^k\}_{k=1}^{K}$. 
Rather than expanding the visual token sequence by concatenating patch tokens, GRF performs token-budget-preserving resolution refinement by injecting localized high-resolution evidence back into the global representation.
For each selected region $\Omega_k$, we extract the corresponding global tokens
$F_g^{\text{roi}} = F_g[\Omega_k]$ and perform a region-conditioned update:
\begin{equation}
\label{eq:grf-update}
\bar{F}_g^{\text{roi}}
=
F_g^{\text{roi}}
+
\mathcal{U}\!\left(F_g^{\text{roi}}, F_p^k\right),
\end{equation}

where $\mathcal{U}(\cdot)$ integrates high-resolution patch features into the aligned global tokens. 
In practice, $\mathcal{U}$ is implemented as a Transformer-style cross-attention block

\begin{equation}
\mathcal{U}(F_g^{\text{roi}},F_p^k)
=
W_O\,\mathrm{Attn}(Q,K,V)
+
\mathrm{FFN}\!\big(\mathrm{LN}(\cdot)\big),
\end{equation}

with $Q = W_Q(F_g^{\text{roi}} + P_Q), K = W_K(F_p^k + P_K), V = W_V(F_p^k + P_K)$.
The refined tokens are written back to their original spatial coordinates
$F_g[\Omega_k] \leftarrow \bar{F}_g^{\text{roi}}$,
yielding the fused representation $F_g^{\text{fused}}$.

\subsection{Enhanced Report Generation}

The fused representation $F_g^{\text{fused}}$ conditions the LLM together with clinical context $C$ to generate the final report:
\begin{equation}
Y^\ast = \argmax_{\hat{Y} \in \mathcal{Y}}
\sum_{t=1}^{T} \log p(\hat{y}_t \mid F_g^{\text{fused}}, C, \hat{y}_{<t}).
\end{equation}

\subsection{Training Objective}

The model is trained end-to-end with a joint objective:
\begin{equation}
\mathcal{L}
=
\mathcal{L}_{\text{RRG}}
+
\lambda \mathcal{L}_{\text{policy}},
\qquad
\mathcal{L}_{\text{RRG}}
=
-\sum_{t=1}^{T}
\log p(y_t \mid F_g^{\text{fused}}, C, y_{<t}).
\label{eq:loss}
\end{equation}
$\mathcal{L}_{\text{policy}}$ provides spatial priors, while $\mathcal{L}_{\text{RRG}}$ supplies task-driven signals that encourage the allocation policy to focus on regions beneficial for report generation.

\section{Experiments}
\label{sec:Experiments}
\subsection{Datasets}

\noindent \textbf{IU-Xray}~\cite{demner2016preparing} contains 7,470 images and 3,955 reports and is a standard benchmark for RRG. Following~\cite{chen2020generating}, we adopt the same 7:1:2 split for training, validation, and testing, and report results on the official test set.

\noindent \textbf{MIMIC-CXR}~\cite{johnson2019mimic} is a large-scale dataset containing 377,110 images and 227,835 reports from 64,588 patients. Following~\cite{chen2020generating}, we use the same split with 270,790 images for training and 3,858 for testing.

\noindent \textbf{CheXpertPlus}~\cite{chambon2024chexpert} is a large-scale benchmark for radiology modeling with 223,228 chest X-rays and 187,711 reports. Following CXPMRG-Bench~\cite{wang2025cxpmrg}, we use the \textit{Findings} section as ground truth and adopt a 7:1:2 split for training, validation, and testing (40,463/5,780/11,562).

\subsection{Experimental Settings}
\noindent\textbf{Evaluation Metrics.}
\textbf{(1) NLG Metrics.}~~Following prior work~\cite{chen2020generating}, we evaluate linguistic quality using BLEU~\cite{papineni2002bleu}, ROUGE-L~\cite{lin2004rouge}, and METEOR~\cite{banerjee2005meteor}.  
\textbf{(2) Clinical Metrics.}~~For clinical correctness, we adopt widely used metrics:
\underline{CheXpert Clinical Efficacy}~\cite{irvin2019chexpert} evaluates alignment between generated findings and ground truth;
\underline{RadGraph~F$_1$}~\cite{jain2021radgraph} measures overlap in entity–relation structures;
\underline{BERTScore}~\cite{zhang2019bertscore} assesses semantic similarity via contextual embeddings;
\underline{RadCliQ}~\cite{yu2023evaluating} combines multiple factors and correlates with radiologist preference.  
\textbf{(3) LLM-based Metrics.}~~We further include two LLM-based metrics for human-aligned factuality:
\underline{GREEN}~\cite{ostmeier2024green} evaluates factual consistency with clinical entities;
\underline{RateScore}~\cite{zhao2024ratescore} measures overall report quality including fluency and diagnostic accuracy.

% % ########### Result Table ###############
\begin{table}[t]
% \small
\centering
\caption{
Comparison on MIMIC-CXR and IU-Xray. Blue shading indicates the \textbf{Top-3} methods: darkest (1st), medium (2nd), light (3rd).
Results marked with $\dagger$ are reported from published literature.
\textit{LLM-based models are labeled with their parameter size.}
}

\resizebox{\linewidth}{!}{
\begin{tabular}{@{}l|l|l|ccccccc@{}}
\hline
\textbf{Dataset} & \textbf{Methods} & \textbf{Publication} & \textbf{BLEU-1} & \textbf{BLEU-2} & \textbf{BLEU-3} & \textbf{BLEU-4} & \textbf{METEOR} & \textbf{ROUGE} \\ \hline

\multirow{18}{*}{\shortstack{\textbf{MIMIC}\\\textbf{CXR}}}
& R2Gen$^{\dagger}$~\cite{chen2020generating} & EMNLP 2020 & 0.353 & 0.218 & 0.145 & 0.103 & 0.142 & - \\
& R2GenCMN$^{\dagger}$~\cite{chen2022cross} & ACL-IJCNLP 2021 & 0.353 & 0.218 & 0.148 & 0.106 & 0.142 & - \\
% & PPKED$^{\dagger}$~\cite{liu2021exploring} & CVPR 2021 & 0.360 & 0.224 & 0.149 & 0.106 & 0.149 & 0.237 \\
& CvT2DistilGPT2$^{\dagger}$~\cite{nicolson2023improving} & AIM 2023 & 0.393 & 0.248 & 0.171 & 0.127 & - & 0.155 \\
& METransformer$^{\dagger}$~\cite{wang2023metransformer} & CVPR 2023 & 0.386 & 0.250 & 0.169 & 0.124 & 0.152 & 0.291 \\
& DCL$^{\dagger}$~\cite{li2023dynamic} & CVPR 2023 & - & - & - & 0.109 & 0.150 & 0.284 \\
& KiUT$^{\dagger}$~\cite{huang2023kiut} & CVPR 2023 & 0.393 & 0.243 & 0.159 & 0.113 & 0.160 & 0.285 \\
& R2GenGPT$^{\dagger}$ (7B)~\cite{wang2023r2gengpt} & Meta-Rad 2023 & 0.411 & 0.267 & 0.186 & 0.134 & 0.160 & 0.297 \\
& EKAGen$^{\dagger}$~\cite{bu2024instance} & CVPR 2024 & 0.419 & 0.258 & 0.170 & 0.119 & 0.157 & 0.287 \\
& Bootstrapping$^{\dagger}$ (14.2B)~\cite{liu2024bootstrapping} & AAAI 2024 & 0.402 & 0.262 & 0.180 & 0.128 & 0.175 & 0.291 \\
& Multi-Grained$^{\dagger}$~\cite{liu2024multi} & TMI 2024 & 0.406 & 0.267 & 0.190 & 0.141 & 0.163 & 0.309 \\
& REVTAF-RRG$^{\dagger}$~\cite{zhou2025learnable} & ICCV 2025 & \cellcolor{top3}0.465 & \cellcolor{top2}0.318 & \cellcolor{top2}0.235 & 0.182 & 0.199 & \cellcolor{top2}0.336 \\
& DAMPER$^{\dagger}$~\cite{huang2025damper} & AAAI 2025 & 0.402 & \cellcolor{top3}0.284 & 0.227 & \cellcolor{top2}0.193 & \cellcolor{top2}0.289 & 0.301 \\
& DATR$^{\dagger}$~\cite{Park_2025_CVPR} & CVPR 2025 & \cellcolor{top2}0.437 & \cellcolor{top3}0.279 & \cellcolor{top3}0.191 & \cellcolor{top3}0.137 & \cellcolor{top3}0.175 & \cellcolor{top3}0.310 \\
& MultiP-R2Gen$^{\dagger}$ (7B)~\cite{chen2025enhancing} & TMI 2025 & 0.425 & 0.279 & 0.194 & 0.140 & 0.167 & 0.307 \\
& KACL$^{\dagger}$ (8B)~\cite{sha2025contrastive} & MICCAI 2025 & 0.414 & 0.270 & 0.184 & 0.136 & 0.169 & 0.303 \\
& MambaXray-VL-Large$^{\dagger}$ (7B)~\cite{wang2025cxpmrg} & CVPR 2025 & 0.422 & 0.268 & 0.184 & 0.133 & 0.167 & 0.289 \\ \cline{2-9}
& \textbf{Ours (4B)} & ECCV 2026 & \cellcolor{top1}\textbf{0.501} & \cellcolor{top1}\textbf{0.376} & \cellcolor{top1}\textbf{0.304} & \cellcolor{top1}\textbf{0.253} & \cellcolor{top1}\textbf{0.336} & \cellcolor{top1}\textbf{0.380} \\ \hline

\multirow{16}{*}{\textbf{IU-Xray}}
& R2Gen$^{\dagger}$~\cite{chen2020generating} & EMNLP 2020 & 0.470 & 0.304 & 0.219 & 0.165 & 0.187 & 0.371 \\
& R2GenCMN$^{\dagger}$~\cite{chen2022cross} & ACL-IJCNLP 2021 & 0.475 & 0.309 & 0.222 & 0.170 & 0.191 & 0.375 \\
% & PPKED$^{\dagger}$~\cite{liu2021exploring} & CVPR 2021 & 0.483 & 0.315 & 0.224 & 0.168 & 0.376 & 0.187 \\
& METransformer$^{\dagger}$~\cite{wang2023metransformer} & CVPR 2023 & 0.483 & 0.322 & 0.228 & 0.172 & 0.192 & 0.380 \\
& KiUT$^{\dagger}$~\cite{huang2023kiut} & CVPR 2023 & \cellcolor{top2}0.525 & 0.360 & 0.251 & 0.185 & \cellcolor{top2}0.242 & \cellcolor{top2}0.409 \\
& R2GenGPT$^{\dagger}$ (7B)~\cite{wang2023r2gengpt} & Meta-Rad 2023 & 0.488 & 0.316 & 0.228 & 0.173 & 0.211 & 0.377 \\
& EKAGen$^{\dagger}$~\cite{bu2024instance} & CVPR 2024 & 0.497 & 0.339 & 0.250 & 0.190 & 0.210 & 0.399 \\
& Bootstrapping$^{\dagger}$ (14.2B)~\cite{liu2024bootstrapping} & AAAI 2024 & 0.499 & 0.323 & 0.238 & 0.184 & 0.208 & 0.390 \\
& Multi-Grained$^{\dagger}$~\cite{liu2024multi} & TMI 2024 & 0.472 & 0.321 & 0.234 & 0.175 & 0.192 & 0.379 \\
& REVTAF-RRG$^{\dagger}$~\cite{zhou2025learnable} & ICCV 2025 & 0.420 & 0.249 & 0.159 & 0.107 & 0.176 & 0.309 \\
& DAMPER$^{\dagger}$~\cite{huang2025damper} & AAAI 2025 & \cellcolor{top3}0.520 & \cellcolor{top2}0.383 & \cellcolor{top2}0.300 & \cellcolor{top2}0.225 & \cellcolor{top3}0.284 & 0.397 \\
& DATR$^{\dagger}$~\cite{Park_2025_CVPR} & CVPR 2025 & 0.486 & \cellcolor{top3}0.348 & \cellcolor{top3}0.265 & \cellcolor{top3}0.208 & 0.205 & \cellcolor{top1}\textbf{0.411} \\
& MultiP-R2Gen$^{\dagger}$ (7B)~\cite{chen2025enhancing} & TMI 2025 & 0.515 & 0.332 & 0.238 & 0.178 & 0.216 & 0.386 \\
& KACL$^{\dagger}$ (8B)~\cite{sha2025contrastive} & MICCAI 2025 & 0.501 & 0.326 & 0.244 & 0.184 & 0.211 & 0.385 \\
& MambaXray-VL-Large$^{\dagger}$ (7B)~\cite{wang2025cxpmrg} & CVPR 2025 & 0.491 & 0.330 & 0.241 & 0.185 & 0.216 & 0.371 \\ \cline{2-9}
& \textbf{Ours (4B)} & ECCV 2026 & \cellcolor{top1}\textbf{0.531} & \cellcolor{top1}\textbf{0.393} & \cellcolor{top1}\textbf{0.324} & \cellcolor{top1}\textbf{0.235} & \cellcolor{top1}\textbf{0.316} & \cellcolor{top3}0.402 \\
\hline
\end{tabular}
}
\label{tab:comparison_sota}
\end{table}

\noindent\textbf{Implementation Details.}
Our model builds on BLIP-3~\cite{xue2024xgen} with a SigLIP vision encoder~\cite{zhai2023sigmoid}, a Perceiver Resampler~\cite{alayrac2022flamingo}, and a Phi-3-mini$_{3.8\text{B}}$ language model~\cite{abdin2024phi}. LSRA is implemented as a lightweight convolutional policy head on the global feature grid (e.g., $27\times27$), using a depthwise $3\times3$ convolution followed by two $1\times1$ convolutions to predict a resolution utility map. During training, Grad-CAM maps from a ResNet-34~\cite{he2016deep,selvaraju2017grad} provide a disease-informed spatial prior to stabilize policy learning (\textbf{not used at inference}). Since BLIP-3 is not designed for medical imaging, all components are fine-tuned: LoRA~\cite{hu2022lora} is applied to the vision encoder and language model (\(r{=}32, \alpha{=}64\)), while the Perceiver Resampler is fully optimized. Training runs for three epochs using AdamW (\(1{\times}10^{-4}\) LR, cosine schedule, warmup ratio 0.03) with batch size 16.

\subsection{Main Results}
We evaluate LePaX on MIMIC-CXR, IU-Xray, and CheXpertPlus using both NLG and clinical metrics. 

\noindent\textbf{NLG Metrics.}~~As shown in Table~\ref{tab:comparison_sota} and Table~\ref{tab:chexpertplus_main}, our method consistently outperforms prior approaches across all datasets and NLG metrics. 
On MIMIC-CXR, it achieves BLEU-4 / METEOR / ROUGE scores of 0.253 / 0.336 / 0.380, surpassing previous encoder–decoder, transformer, and LLM-based methods. 
On IU-Xray, it reaches 0.235 / 0.316 / 0.402, yielding an average 11\% improvement over the previous best. 
On CheXpertPlus, our model also achieves state-of-the-art performance on both NLG and clinical efficacy metrics, demonstrating strong generalization.
Together, these results indicate that our high-resolution patch-guided representation captures subtle disease cues while maintaining coherent global–local semantics, leading to accurate and clinically faithful reports.

% ########### Chexpert Plus ###############
\begin{table}[t]
\centering
\caption{Experimental results on the CXPMRG-Bench~\cite{wang2025cxpmrg}. NLG metrics: BLEU-4, ROUGE, METEOR. CE metrics: Precision, Recall, and F1.}
\label{tab:chexpertplus_main}
\setlength{\tabcolsep}{5pt}
\resizebox{\linewidth}{!}{
\begin{tabular}{l|l|ccc|ccc}
\hline
\textbf{Methods} & \textbf{Publish} & \textbf{BLEU-4} & \textbf{ROUGE} & \textbf{METEOR} & \textbf{Precision} & \textbf{Recall} & \textbf{F1} \\
\hline
ORGan~\cite{hou2023organ}          & ACL 2023      & 0.086 & 0.261 & 0.135 & 0.288 & 0.287 & 0.277 \\
M2KT~\cite{yang2023radiology}      & MIA 2021      & 0.078 & 0.247 & 0.101 & 0.044 & 0.142 & 0.058 \\
TIMER~\cite{wu2023token}           & CHIL 2023     & 0.083 & 0.254 & 0.121 & 0.345 & 0.238 & 0.234 \\
CvT2DistilGPT2~\cite{nicolson2023improving} & AIM 2023 & 0.067 & 0.238 & 0.118 & 0.285 & 0.252 & 0.246 \\
R2Gen~\cite{chen2020generating}    & EMNLP 2020    & 0.081 & 0.246 & 0.113 & 0.318 & 0.200 & 0.181 \\
R2GenCMN~\cite{chen2022cross}      & ACL 2021      & 0.087 & 0.256 & 0.127 & 0.329 & 0.241 & 0.231 \\
Zhu~et~al.~\cite{zhu2023utilizing} & MICCAI 2023   & 0.074 & 0.235 & 0.128 & 0.217 & 0.308 & 0.205 \\
CAMANet~\cite{wang2024camanet}     & IEEE JBHI 2023& 0.083 & 0.249 & 0.118 & 0.328 & 0.224 & 0.216 \\
Token-Mixer~\cite{yang2024token}   & IEEE TMI 2023 & 0.091 & 0.261 & 0.135 & 0.309 & 0.270 & 0.288 \\
PromptMRG~\cite{jin2024promptmrg}  & AAAI 2024     & 0.095 & 0.222 & 0.121 & 0.258 & 0.265 & 0.281 \\
R2GenGPT~\cite{wang2023r2gengpt}   & Meta-Rad.2023& 0.101 & 0.266 & 0.145 & 0.315 & 0.244 & 0.260 \\
R2GenCSR~\cite{wang2024r2gencsr}   & arXiv 2024    & 0.100 & 0.265 & 0.146 & 0.315 & 0.247 & 0.259 \\
MambaXray-VL-B~\cite{wang2025cxpmrg} & CVPR 2025     & \cellcolor{top3}0.105 & \cellcolor{top2}0.267 & \cellcolor{top3}0.149 & \cellcolor{top3}0.333 & \cellcolor{top3}0.264 & \cellcolor{top3}0.273 \\
MambaXray-VL-L~\cite{wang2025cxpmrg} & CVPR 2025     & \cellcolor{top2}0.112 & \cellcolor{top1}\textbf{0.276} & \cellcolor{top2}0.157 & \cellcolor{top2}0.377 & \cellcolor{top2}0.319 & \cellcolor{top2}0.335 \\
\hline
\textbf{Ours} & ECCV 2026 & \cellcolor{top1}\textbf{0.138} & \cellcolor{top3}0.252 & \cellcolor{top1}\textbf{0.227} & \cellcolor{top1}\textbf{0.420} & \cellcolor{top1}\textbf{0.368} & \cellcolor{top1}\textbf{0.363} \\
\hline
\end{tabular}
}
\end{table}

\noindent\textbf{Clinical Metrics.}~~
As shown in Table~\ref{Table:ComparisonWithSOTA_Rad} and Table~\ref{tab:clinic_metrics}, our model achieves strong clinical fidelity across evaluation metrics. Since large-scale expert review of thousands of reports is impractical, we adopt RadCliQ and GREEN, human-aligned metrics trained on radiologist ratings that correlate strongly with expert judgments. On MIMIC-CXR, our model achieves the highest RadGraph~F$_1$ (0.346) and BERTScore (0.526), together with a lower RadCliQ (0.665$\downarrow$), indicating closer alignment with expert reports. It also obtains 0.413 GREEN and 0.627 RateScore on LLM-based evaluations. Under the CheXpert Clinical Efficacy framework, the model reaches 0.595 precision, 0.479 recall, and 0.531 F$_1$, surpassing recent transformer- and LLM-based baselines.

% ########### Radcliq, green, ratescore ###############
\begin{table}[h]
\caption{
Comparison with state-of-the-art methods on MIMIC-CXR.
RG$_{F1}$: RadGraph F1; 
BERT: BERTScore; 
CliQ: RadCliQ; 
GREEN: clinical consistency score; 
Rate: human rating score.
}
\centering
\resizebox{\linewidth}{!}{
\begin{tabular}{l|l|ccccc}
\hline
\textbf{Methods} & \textbf{Publication} 
& \textbf{RG$_{F1}$($\uparrow$)} 
& \textbf{BERT($\uparrow$)} 
& \textbf{CliQ($\downarrow$)} 
& \textbf{GREEN($\uparrow$)} 
& \textbf{Rate($\uparrow$)} \\
\hline

R2Gen~\cite{chen2020generating} 
& EMNLP 2020 
& 0.172 & 0.406 & 1.228 & 0.276 & 0.526 \\

% R2GenCMN~\cite{chen2022cross} 
% & ACL-IJCNLP 2021 
% & 0.182 & 0.418 & 1.182 & 0.297 & \cellcolor{top2}0.538 \\

CvT2DistilGPT2~\cite{nicolson2023improving} 
& AI in Med 2023 
& 0.196 & 0.374 & 1.220 & \cellcolor{top2}0.320 & 0.527 \\

RaDialog-RG~\cite{pellegrini2023radialog} 
& MIDL 2024 
& - & 0.400 & - & - & - \\

PromptMRG~\cite{jin2024promptmrg} 
& AAAI 2024 
& 0.190 & 0.357 & 1.169 & 0.287 & \cellcolor{top3}0.528 \\

R2GenGPT~\cite{wang2023r2gengpt} 
& Meta-Rad 2023 
& 0.187 & 0.415 & 1.207 & 0.300 & \cellcolor{top3}0.528 \\

KARGEN~\cite{li2024kargen} 
& MICCAI 2024 
& \cellcolor{top3}0.203 
& \cellcolor{top3}0.421 
& 1.165 
& \cellcolor{top3}0.308 
& \cellcolor{top2}0.533 \\

EKAGen~\cite{bu2024instance} 
& CVPR 2024 
& 0.199 & 0.412 & \cellcolor{top2}1.126 & 0.256 & 0.512 \\

MultiP-R2Gen~\cite{chen2025enhancing} 
& TMI 2025 
& \cellcolor{top2}0.208 
& \cellcolor{top2}0.425 
& \cellcolor{top3}1.140 
& - & - \\

\hline
\textbf{Ours} 
& ECCV 2026 
& \cellcolor{top1}\textbf{0.346} 
& \cellcolor{top1}\textbf{0.526} 
& \cellcolor{top1}\textbf{0.665} 
& \cellcolor{top1}\textbf{0.413} 
& \cellcolor{top1}\textbf{0.627} \\
\hline

\end{tabular}
}
\label{Table:ComparisonWithSOTA_Rad}
\end{table}
% #########################

% ########### CheXpert clinical efficacy ###############
\begin{table}[t]
\centering
\caption{Evaluation of CheXpert clinical efficacy metrics on the MIMIC-CXR dataset.}
\label{tab:clinic_metrics}
\resizebox{0.7\linewidth}{!}{
\begin{tabular}{l|l|ccc}
\hline
\textbf{Methods} & \textbf{Publication} & \textbf{Precision} & \textbf{Recall} & \textbf{F1} \\
\hline

R2Gen~\cite{chen2020generating} & EMNLP 2020 & 0.333 & 0.273 & 0.276 \\
% R2GenCMN~\cite{chen2022cross} & ACL-IJCNLP 2021 & 0.334 & 0.275 & 0.278 \\
R2GenGPT~\cite{wang2023r2gengpt} & Meta-Rad 2023 & 0.392 & 0.387 & 0.389 \\
METransformer~\cite{wang2023metransformer} & CVPR 2023 & 0.364 & 0.309 & 0.311 \\
Multi-Grained~\cite{liu2024multi} & TMI 2024 & 0.457 & 0.337 & 0.330 \\
MambaXray-VL-Large~\cite{wang2025cxpmrg} & CVPR 2025 & 0.371 & 0.321 & 0.340 \\
KACL~\cite{sha2025contrastive} & MICCAI 2025 & 0.503 & 0.442 & 0.469 \\

DAMPER~\cite{huang2025damper} & AAAI 2025 
& \cellcolor{top3}0.512 
& \cellcolor{top3}0.473 
& \cellcolor{top3}0.507 \\

DART~\cite{Park_2025_CVPR} & CVPR 2025 
& \cellcolor{top2}0.520 
& \cellcolor{top1}\textbf{0.546} 
& \cellcolor{top1}\textbf{0.533} \\

\hline
\textbf{Ours} & ECCV 2026 
& \cellcolor{top1}\textbf{0.595} 
& \cellcolor{top2}0.479 
& \cellcolor{top2}0.531 \\
\hline

\end{tabular}
}
\end{table}

\subsection{Ablation Study}

%%%%%%%%%%%%%%%%%%ablation mimic component%%%%%%%%%%%%%%%

%%%%%%%%%%%%%%%%%%%%%%%%%%%%%%%%%
\noindent\textbf{Contribution of Each Component.}~~
We perform ablation studies on MIMIC-CXR (Table~\ref{tab:ablation_mimic}) to evaluate each component. The baseline without either module achieves the lowest performance, indicating limited ability to capture fine-grained pathological cues.

\noindent\textbf{GRF.}~~
Introducing GRF with randomly sampled patches already improves performance (BLEU-4: +0.006, METEOR: +0.013, ROUGE: +0.010) under the same patch budget as LSRA, indicating that regional high-resolution evidence becomes beneficial when fused with global representations.

\noindent\textbf{LSRA.}~~
Replacing random sampling with guided region selection further improves performance. Grad-CAM increases B-4 to 0.251, while LSRA achieves the best result (B-4: 0.253) under the same token budget (128) and comparable inference time, demonstrating the advantage of task-driven resolution allocation.

\noindent\textbf{Policy supervision.}~~
Training LSRA with report-only supervision leads to less stable optimization, as the report objective provides indirect signals for discrete region allocation. Adding policy supervision improves performance, indicating that report supervision and spatial priors provide complementary guidance.

\noindent\textbf{Fusion strategy.}~~
Simply concatenating tokens without GRF degrades performance and increases the input sequence length (e.g., $(1_{\text{global}} + 4_{\text{patches}})\times128 = 640$ tokens), resulting in much higher inference cost (6.5s per case).

\noindent\textbf{Full model.}~~
The complete model maintains a fixed token budget (128 tokens) while enriching token semantics through GRF achieving the best performance with only a small inference overhead (2.7s per case). As shown in Fig.~\ref{fig:info}, GRF increases feature variance and L2 norm, indicating that high-resolution regional evidence enhances token expressiveness without increasing token count.

Overall, these results highlight the complementary roles of LSRA and GRF: LSRA adaptively allocates resolution to informative regions, while GRF integrates regional evidence into global representation in a token-efficient manner.

\begin{figure}[t]
  \centering
    \includegraphics[width=\linewidth]{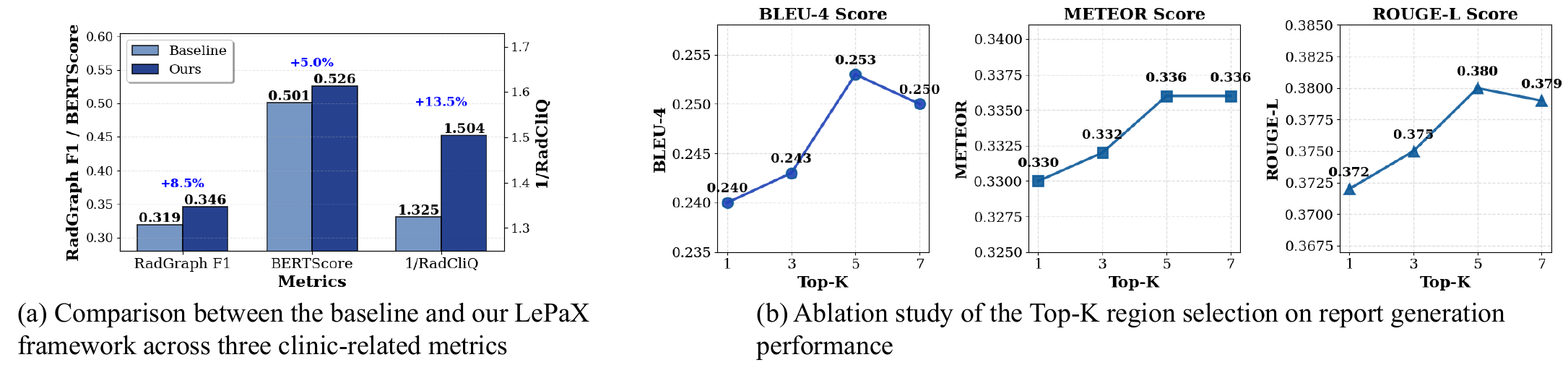}
    \caption{Ablation studies on MIMIC-CXR.}
    \label{fig:clinic-ablation}
\end{figure}

\noindent\textbf{Clinical Evaluation.}~~
We further evaluate clinical reliability using RadGraph~F$_1$, BERTScore, and RadCliQ (Fig.~\ref{fig:clinic-ablation}(a)).
\textbf{LePaX} improves all metrics, achieving relative gains of \textbf{+8.5\%} in RadGraph~F$_1$, \textbf{+5.0\%} in BERTScore, and \textbf{+13.5\%} in $1/\mathrm{RadCliQ}$.
These results indicate improved factual correctness and clinical consistency in the generated reports.

\begin{table}[t]
\centering
\caption{
Ablation study on MIMIC-CXR.
GRF: Global--Regional Fusion; Grad-CAM: external localization-based patch selection; LSRA: Learnable Spatial Resolution Allocation.
\textbf{LSRA Loss}: report-only supervision vs.\ report + policy supervision.
}
\label{tab:ablation_mimic}
\resizebox{\linewidth}{!}{
\begin{tabular}{ccc|ccc|c|c}
\hline
\multicolumn{3}{c|}{\textbf{Modules}} 
& \multicolumn{3}{c|}{\textbf{NLG Metrics}} 
& \multirow{2}{*}{\textbf{\#Vision Tokens}} 
& \multirow{2}{*}{\textbf{Inference Time (case)}} \\
\cline{1-6}
GRF & Grad-CAM & LSRA 
& B-4 & METEOR & ROUGE 
&  &  \\
\hline
-- & -- & -- 
& 0.237 & 0.313 & 0.350 
& 128 & 2.3s \\

\checkmark & -- & -- 
& 0.243 & 0.326 & 0.360 
& 128 & 2.6s \\

-- & \checkmark & -- 
& 0.234 & 0.318 & 0.334 
& $\sim$640 & 6.5s \\

-- & -- & \checkmark 
& 0.235 & 0.318 & 0.333 
& $\sim$640 & 6.5s \\

\checkmark & \checkmark & -- 
& 0.251 & 0.334 & 0.379 
& 128 & 2.6s \\

\checkmark & -- & \checkmark\,(report-only) 
& 0.246 & 0.329 & 0.365 
& 128 & 2.7s \\

\checkmark & -- & \checkmark\,(report+policy) 
& \textbf{0.253} & \textbf{0.336} & \textbf{0.380} 
& 128 & 2.7s \\
\hline
\end{tabular}}
\end{table}

\noindent\textbf{Resolution Allocation Efficiency.}~~
We analyze the impact of input resolution on report generation (Table~\ref{tab:resolution}). Increasing resolution from $384$ to $1024$ yields only marginal gains while greatly increasing cost, with vision tokens rising from 128 to 1280 and inference time reaching 10.0s per case. In contrast, \textbf{LePaX} allocates high-resolution capacity to informative regions under a fixed token budget. Even with native-resolution images (up to $1920\times1920$), it achieves better performance with only a small increase in inference time, demonstrating greater efficiency than uniform resolution scaling.
\begin{table}[t]
\centering
\caption{
Comparison of uniform resolution scaling and LePaX. Selective high-resolution allocation achieves better report quality under a fixed visual token budget with significantly lower computational cost.
}
\resizebox{\linewidth}{!}{
\begin{tabular}{c|ccccc|ccc}
\hline
\multicolumn{1}{c|}{\textbf{Setting}} 
& \multicolumn{5}{c|}{\textbf{Report Quality Metrics}}
& \multicolumn{3}{c}{\textbf{Efficiency}} \\

\cline{2-9}
 & B-4 & ROUGE & RG$_{F1}$ & BERT & 1/CliQ
 & Vision Tokens & Time/Case & GPU Mem \\

\hline

Uniform-384
& 0.237 & 0.350 & 0.319 & 0.501 & 1.325
& 128 & 2.3s & 36GB \\

Uniform-1024
& 0.240 & 0.352 & 0.320 & 0.504 & 1.329
& 1280 & 10.0s & 65GB \\

\hline

Ours-1024
& 0.245 & 0.364 & 0.331 & 0.513 & 1.389
& 128 & 2.4s & 40GB \\

\textbf{Ours-1920}
& \textbf{0.253} & \textbf{0.380} & \textbf{0.346} & \textbf{0.526} & \textbf{1.504}
& \textbf{128} & \textbf{2.7s} & \textbf{46GB} \\

\hline
\end{tabular}
}
\label{tab:resolution}
\end{table}

\noindent\textbf{Ablation on Top-K Region Selection.}~~We evaluate different Top-K values (1, 3, 5, 7) for high-resolution patch extraction (Fig.~\ref{fig:clinic-ablation}(b)). Increasing K improves performance by incorporating more disease-relevant regions, while overly large K introduces redundancy and slightly degrades report quality. \textbf{Top-K=5} achieves the best results (B-4: 0.253, METEOR: 0.336, ROUGE-L: 0.380).

\begin{figure}[t]
  \centering
   \includegraphics[width=\linewidth]{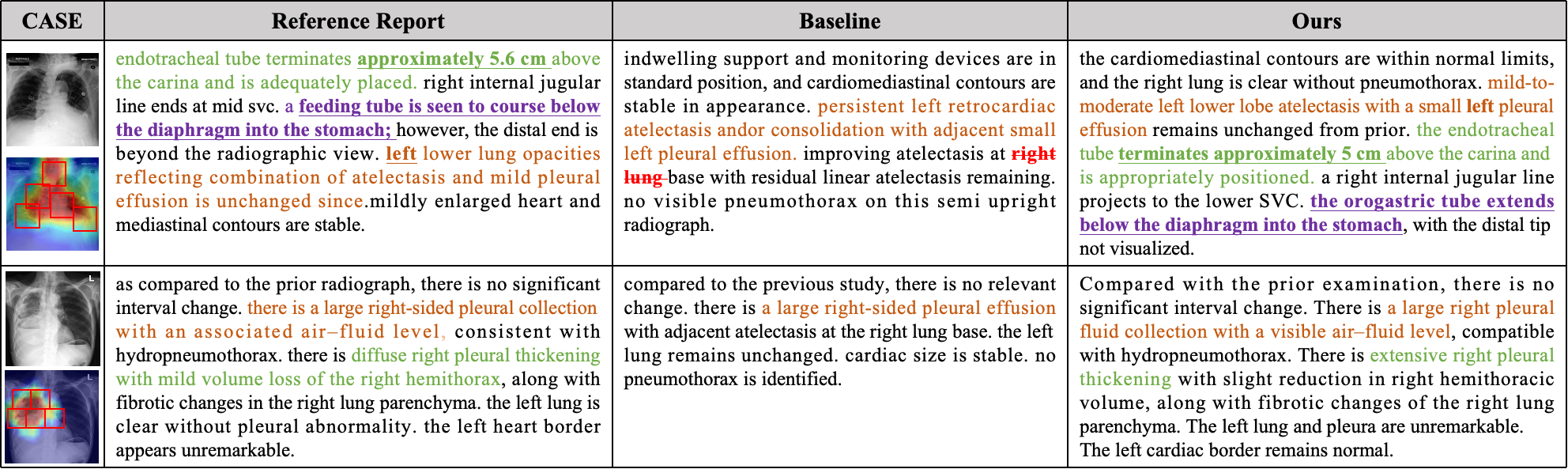}
   \caption{Qualitative comparison between the baseline and our method on two MIMIC-CXR cases. The main clinical findings are highlighted in different colors for clarity.}
   \label{fig:qualitative}
\end{figure}

\subsection{Qualitative Analysis}
\begin{figure}[h]
\centering
\begin{subfigure}{0.3\linewidth}
    \centering
    \includegraphics[width=\linewidth]{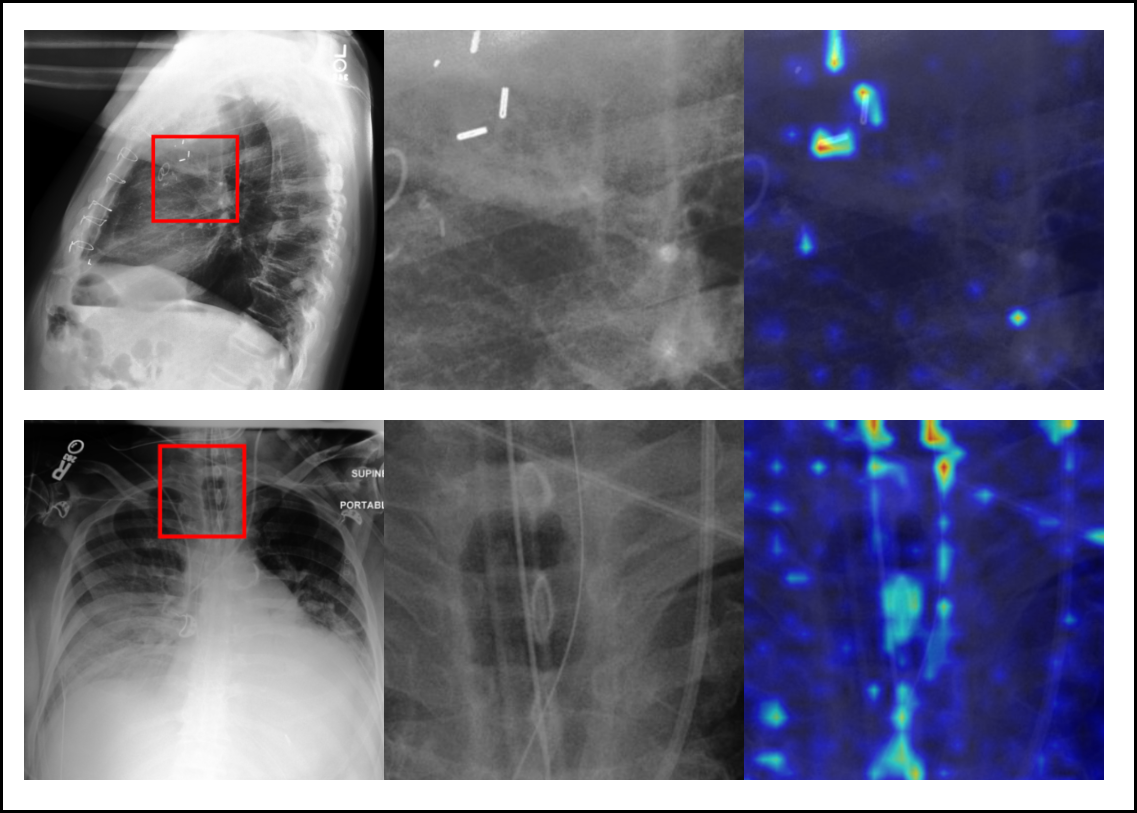}
    \caption{Adaptive region selection and LR$\rightarrow$HR fusion}
    \label{fig:attention}
\end{subfigure}
\hspace{0.05\linewidth}
\begin{subfigure}{0.32\linewidth}
    \centering
    \includegraphics[width=\linewidth]{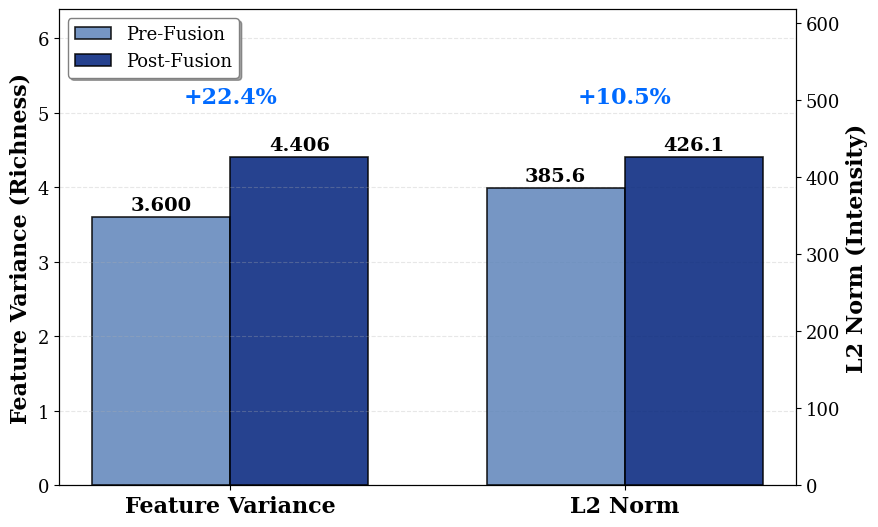}
    \caption{Feature statistics before and after GRF}
    \label{fig:info}
\end{subfigure}

\caption{Analysis of the proposed Global–Regional Fusion (GRF). (a) Visualization of selected regions and LR$\rightarrow$HR fusion attention. (b) Feature statistics before and after fusion, showing increased variance and L2 norm after GRF.}
\label{fig:grf_analysis}
\end{figure}

Fig.~\ref{fig:qualitative} compares the baseline and \textbf{LePaX} on MIMIC-CXR cases. Grad-CAM heatmaps show that diagnostically relevant regions occupy only a small portion of the image, motivating selective high-resolution patch extraction. While the baseline detects findings such as \textit{pleural effusion} and \textit{tube placement}, its descriptions are often coarse or incomplete. In contrast, \textbf{LePaX} integrates lesion-focused details with global context to produce more precise and clinically grounded reports. For example, in \textbf{Case~I}, the model localizes the \textit{endotracheal tube} relative to the carina and identifies the \textit{orogastric tube below the diaphragm}, consistent with the reference report. Fig.\ref{fig:attention} further visualizes selected low-resolution (LR) regions, their corresponding high-resolution (HR) patches, and LR$\rightarrow$HR fusion attention maps. Red boxes denote LR regions selected for high-resolution processing. In \textbf{Case~I}, attention highlights subtle post-surgical structures such as surgical clips, while in \textbf{Case~II} it follows the course of the right internal jugular line in the upper mediastinum. These examples show how GRF refines coarse spatial cues and enhances diagnostically relevant regions, enabling spatially grounded reasoning beyond classifier-driven activation.

\begin{figure}[h]
  \centering
   \includegraphics[width=0.8\linewidth]{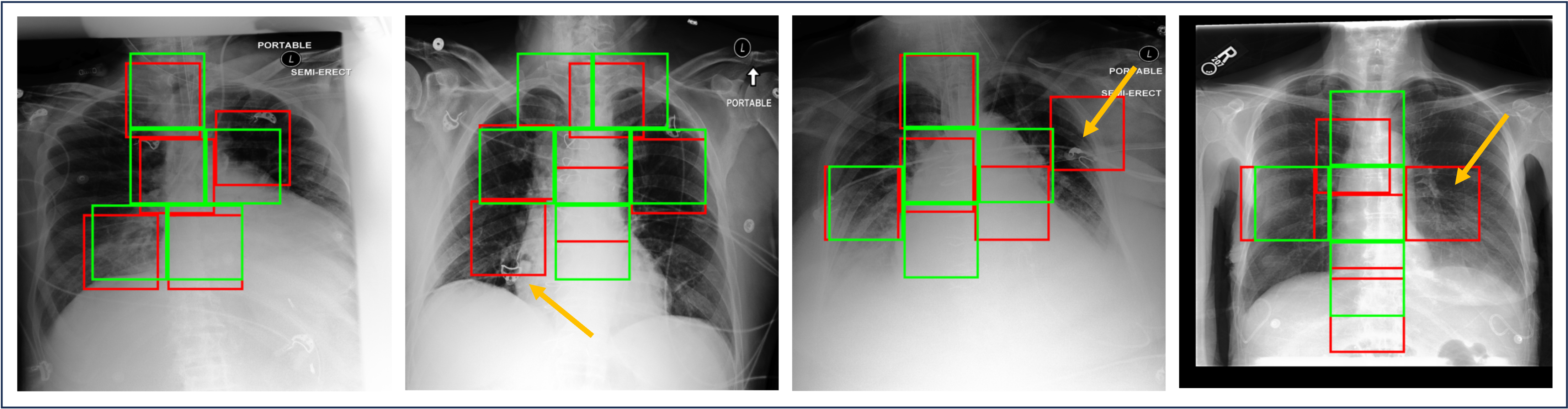}
   \caption{Qualitative comparison of region allocation. Red boxes denote LSRA predictions and green boxes indicate Grad-CAM guidance. LSRA consistently identifies clinically relevant regions similar to Grad-CAM. Cases~2–4 further show that report-level supervision helps the learned policy better localize medical devices (Cases~2–3) and pleural effusion (Case~4).}
   \label{fig:camcase}
\end{figure}

Additionally, Fig.~\ref{fig:camcase} compares LSRA and Grad-CAM region allocation. Red boxes denote LSRA predictions and green boxes indicate Grad-CAM guidance. Across the examples, LSRA consistently captures clinically relevant regions highlighted by Grad-CAM. Cases~2–4 further demonstrate the effect of report-level supervision, where the learned policy better localizes medical devices (Cases~2–3) and pleural effusion (Case~4). These results show that LSRA preserves diagnostically important regions while adapting toward report-oriented spatial relevance.

% \begin{figure}[h]
%   \centering
%    \includegraphics[width=\linewidth]{attention.png}
%    \caption{Visualization of adaptive region selection and high-resolution fusion. For each case, we show the selected low-resolution (LR) region (red box), the corresponding high-resolution (HR) patch, and the LR$\rightarrow$HR fusion attention map.}
%    \label{fig:attention}
% \end{figure}

\section{Conclusions}
\label{sec:Conclusions}
We propose LePaX, a high-resolution RRG framework with two complementary components: LSRA for lesion-focused patch extraction and GRF for hierarchical fusion of regional and global cues. This shifts RRG from coarse, pixel-limited encoding toward clinically guided high-resolution understanding. By combining lesion-focused precision with context-aware interpretation, LePaX achieves consistent gains over prior methods on both lexical and clinical metrics across MIMIC-CXR, IU-Xray, and CheXpertPlus.

\newpage
% \section*{Acknowledgements}
% Please insert your acknowledgments here.

\bibliographystyle{splncs04}
\bibliography{main}
\end{document}